# The Cerebellum: New Computational Model that Reveals its Primary Function to Calculate Multibody Dynamics Conform to Lagrange-Euler Formulation

**Lavdim Kurtaj[1], Ilir Limani[2], Vjosa Shatri[3] and Avni Skeja[4]**

**[1] Department of Automation, Faculty of Electrical and Computer Engineering, University of Prishtina
Prishtina, 10000, Kosovo**

**[2] Department of Automation, Faculty of Electrical and Computer Engineering, University of Prishtina
Prishtina, 10000, Kosovo**

**[3] Department of Fundamental Engineering Subjects, Faculty of Electrical and Computer Engineering,
University of Prishtina
Prishtina, 10000, Kosovo**

**[4] Department of Automation, Faculty of Electrical and Computer Engineering, University of Prishtina
Prishtina, 10000, Kosovo**

## Abstract

Cerebellum is part of the brain that occupies only 10% of the brain volume, but it contains about 80% of total number of brain neurons. New cerebellar function model is developed that sets cerebellar circuits in context of multibody dynamics model computations, as important step in controlling balance and movement coordination, functions performed by two oldest parts of the cerebellum. Model gives new functional interpretation for granule cells-Golgi cell circuit, including distinct function for upper and lower Golgi cell dendrite trees, and resolves issue of sharing Granule cells between Purkinje cells. Sets new function for basket cells, and for stellate cells according to position in molecular layer. New model enables easily and direct integration of sensory information from vestibular system and cutaneous mechanoreceptors, for balance, movement and interaction with environments. Model gives explanation of Purkinje cells convergence on deep-cerebellar nuclei.

## 1. Introduction

The cerebellum is part of vertebrate brain, and it is only part that spans to both sides continually without interruption. It plays important role in lower levels functions of motor control in relation to sensory information, such as coordination of complicated multi-joint bodies, balance, motor planning. According to connections with other parts of the brain, it is suggested for cerebellum to be involved in many other functional levels, up to cognition and emotional level. What makes debatable broad and diverse functional involvement of cerebellum is its uniform structure, in neuronal elements present, in local connectivity between them, and in circuits with other parts of the brain. This has lead authors of [1] to propose the general-purpose coprocessor function for the cerebellum. Based on connections between neuronal elements cerebellum can be viewed as a hierarchically organized system. Lowest level contains array of elementary processing units, each with group of neurons and single Purkinje cell, all arranged in highly ordered fashion [2]. Starting from functions of oldest parts of the cerebellum, and importance of robot dynamics model for high performance control algorithms, new cerebellar function model is developed that relates cerebellar circuits with multibody multijoint dynamics model computations.

## 2. Multibody System Dynamics

Meaning of term multibody system is very broad. In following, this term will imply to a set of bodies, called links, assembled together in a specific way with a set of movement constraining elements, called joints [3]. Interaction between bodies can happen only in joints. Joint can be only a concept that describes relative position between bodies, or specific construction where bodies are rigidly connected. Former case of joint infers that constraint motion is achieved with special construction of link endings that will be assembled together. In later case, all bodies that are rigidly connected together are treated as single link. Minimum number of parameters required to define relative movement of links connected with a joint denotes joint degrees of freedom. Sum of all joint degrees of freedom gives multibody system degrees of freedom. Each joint may have actuator for intentional altering of joint position (active joint), otherwise joint is passive and its state may change from external interactions acting on corresponding links. Links may be rigid or flexible. In general all links will express some degree of flexibility (elasticity), but for normal operating conditions deformation are bellow some maximal acceptable value. Theory of multibody system can be used to analyze all animals with skeletal structure (vertebrata), including humans, were links are represented by bones and joints are usually called articulations [4]. Muscles are used for joints actuation. A typical artificial multibody system, that sometimes serves as synonym for them, are robots [5], [6], sometimes with structure inspired by nature.

Robotics is interdisciplinary field and many disciplines contribute to different aspects of its development, naming some: mechanics, electronics, mechatronics, control engineering, computer science, biomechanics, etc. Main topics of robotics are kinematics, dynamics, trajectory planning, sensing, motion control, and intelligence.

## 2.1 Robot kinematics

Robot consists of links connected with joints. Robot kinematics deals with problem of finding analytical model that describes robot motion, with respect to fixed coordinate frame over time, without regard to forces and/or moments that cause the motion. Forward kinematics will give position $p$ and orientation $\psi$ of some frame attached to the robot, usually end-effector frame, as a function of joint variables $q$

$$\begin{bmatrix} p \\ \psi \end{bmatrix} = f_{FK}(q).$$  (1)

Opposite problem of finding joint variables $q$ for given end-effector position $p$ and orientation $\psi$ is solved with inverse kinematics. Solution of forward kinematics is unique, whereas solution of inverse kinematics is not, necessitating sometime to add additional variables [7] and sometime to use numerical methods [8].

## 2.2 Robot dynamics

General form of dynamics equation of motion in matrix form (inverse dynamics) for robots with rigid links, derived following Lagrange-Euler formulation (L-E) is

$$\tau = D(q)\ddot{q} + C(q,\dot{q}) + G(q) + F_f(\dot{q}) + J^T(q)f_e , \qquad (2)$$

and gives torques/forces $\tau$ needed to drive joints for desired trajectory given by joint variables $q$, $\dot{q}$, and $\ddot{q}$, for joint position, velocity, and acceleration.

$D(q)$ denotes inertia matrix, $C(q,\dot{q})$ is the vector of Coriolis and centripetal terms, $G(q)$ is the vector of gravity terms. Friction is represented with $F_f(\dot{q})$ and any friction model may be used. Usually it includes dynamic friction and other unstructured friction effects. Possible interaction of end-effector with environment is expressed with last term, where $f_e$ is vector of external forces and torques, and $J(q)$ is robot Jacobian. Eq. (2) is used for different forms of control strategies, e.g., feedforward, computed torque, feedback linearization [9], [10].

Forward dynamics equation can be derived from (2) by solving it for acceleration

$$\ddot{q} = D^{-1}(q)\left(\tau - C(q,\dot{q}) - G(q) - F_f(\dot{q}) - J^T(q)f_e\right), \qquad (3)$$

and is used for simulation of robot movement.

## 2.3 Processing units for robot dynamics computation

Advanced control strategies rely on dynamics model of the robot to achieve and maintain desired system dynamic response in accordance with specified performance criteria. To be able to track desired trajectories as close as possible, computation of appropriate torques/forces to drive joint actuators of the robot has to be done in real time. Delay caused by computer system to calculate control algorithm has negative influence on system performance. To shorten this delay when control computer system is based on single processor, computation is based on Newton-Euler formulation (N-E) for dynamics model, as a set of forward and backward recursive equations. Computational complexity of N-E is of order O(n), where n is the number of degrees of freedom (DOF) of the robot, as compared to O(n^4) for L-E. Two main additional approaches are recursive Lagrangian formulation [11] with O(n) complexity, and generalized d'Alambert equations of motion [12] with complexity of order O(n^3).

Classification of computational complexity is based in number of multiplications and additions for dynamics model calculation ($n_{dm}$). When calculations are done in single processor computers with given execution time per instruction ($T_{ins}$), total number of instructions $n_{dm}$ will determine control system time delay introduced to control system by computation of dynamics model to $T_{sp}=n_{dm}\cdot T_{ins}$. For computer architectures with many processors, recursive computations will limit minimum delay proportional to the number of recursion steps ($n_{rs}$), i.e. $T_{mpr}=n_{rs}\cdot T_{ins}$, even under assumption that calculation of single recursion step is one "super" instruction and executes within $T_{ins}$. This can not be made smaller because of data dependencies between recursion steps. For N-E case there are n forward and n backward steps and $n_{rs}=2 \cdot n$, where n is number of degrees of freedom for the robot.

If for some multibody system, with n=10 and with many very slow processors with $T_{ins}=100\mu s$, control system must update control values every $T_c=10ms$, question is: what the control computer architecture should be and which set of "super" instructions should it posses? Simple solution that pops up, when it is required to do complex, time consuming calculations with slow processors is to store precalculated values in a table. "Super" instruction would be the one that reads input data, calculates address from input data, reads value from the table, and writes result to the output. If input data are directly used as address, processor would be nothing but a memory of sufficient capacity. Calculation time with this "computer" system would be $1 \cdot T_{ins}$. Despite its simplicity, this approach has as a drawback enormous size of memory for most applications. Input data space dimension for robot with n DOF is 3n (n joint positions, n velocities, and n accelerations). If each dimension of input data space is quantized with $b_k$ values, memory size would be $b_1 \cdot b_2 \cdots b_{3n}$, indicating exponential growth. If all $b_k$ are equal memory size would be $b_k^{3n}$, and for $b_k=16$ (that is considered to be very coarse quantization) it would be $16^{30}$, i.e. 120-bit address and $1024^{12}$ byte memory space, if single byte values are stored. Decreasing memory size of this unstructured approach is sought by exploiting structure of the problem to be solved.

Dynamics equation of motion for single joint is

$$\tau_k = d_{k1}(q)q_1 + d_{k2}(q)q_2 + \quad + d_{kn}(q)q_n +$$

$$\begin{bmatrix} h_{k11}(q)q_1 + h_{k21}(q)q_2 + & + h_{kn1}(q)q_n \\ h_{k12}(q)q_1 + h_{k22}(q)q_2 + & + h_{kn2}(q)q_n \\ \\ h_{k1n}(q)q_1 + h_{k2n}(q)q_2 + & + h_{knn}(q)q_n \end{bmatrix}^T \begin{bmatrix} q_1 \\ q_2 \\ \\ q_n \end{bmatrix} +$$

$$G_{k1}(q)g_{x_0} + G_{k2}(q)g_{y_0} + G_{k3}(q)g_{z_0} +$$

$$F_{kd}q_k + F_{ks}(q_k) +$$

$$J_{1k}(q)F_{ex_0} + J_{2k}(q)F_{ey_0} + J_{3k}(q)F_{ez_0} +$$

$$J_{4k}(q)M_{ex_0} + J_{5k}(q)M_{ey_0} + J_{6k}(q)M_{ez_0}$$

$$(4)$$

where gravity term for joint k, $G_k(q)$, has been written in form so calculations with individual components of gravitational constant vector $g$ are visible explicitly. This form has two advantages; first, it creates gravity terms that are similar in form with most of others, and second, it opens possibility to compensate directly when multibody reference changes orientation relative to gravitational field, that is present almost all the time for example in humans and mobile robots.

To calculate dynamics model for single joint, as seen from Eq. (4), we need only a small set of prototype functions. Main function prototype of form

$$f_p(q,x) = N(q)x ,$$ (5)

will suffice for calculation of inertial torques, $d_{kj}(q)\ddot{q}_j$, gravitational torques, $G_{kj}(q)g_j$, and torques exerted from interaction with environment, $J_{jk}(q)F_{ej}$ and $J_{jk}(q)M_{ej}$. Calculation of Coriolis and centrifugal torques $h_{kij}(q)\dot{q}_i\dot{q}_j$ can be done by using first prototype function

$$\tau_{Cc} = f_p(q,x)y ,$$ (6)

or they can be calculated by defining second prototype function

$$f_{p2}(q,x,y) = N(q)xy .$$ (7)

Calculating previous terms of dynamics equation with second prototype can be done by giving constant value equal to 1 for second variable, $y$=1, or by providing some controlled switching mechanism for variable $y$. Calculation of dynamic and static friction torques, $F_{kd}\dot{q}_k$ and $F_{ks}(\dot{q}_k)$, is possible with either prototype function, but, having in mind their simplicity compared to other terms, it seams like waste of computational resources. This could be solved with some add-on to selected prototype function. Extended second prototype function would become

$$f_{p2s+}(q,x,s,y,a,b) = N(q)xy(s) + \mu_d a + \mu_s (b) ,$$ (8)

were we have included logical variable $s$ as switch for variable $y$, with $y(s)$=1 for one value of s and $y(s)$=$y$ for the other. Two last terms of Eq. (8) can be used to calculate friction torques of Eq. (4).

Computer architecture for calculation of dynamics model that exploits structure of the model would have first layer with one processing unit (PU) for each term in Eq. (4) for each DOF, excluding friction terms, with single "super" instruction for doing calculation according to Eq. (8) within $T_{ins}$ time, and second layer with one PU for each DOF for summing corresponding results together, also within $T_{ins}$ time. Total calculation time would be $2 \cdot T_{ins}$, independent of DOF for the system. First layer of architecture resembles Single Instruction Multiple Data (SIMD) class of parallel computer architectures according to Flynn's taxonomy [13], [14]. Number of PU in first layer for n=10 is n·(n+n·n+3+3+6)=1220, and second layer has 10 PU. With new structure, dimension of input space for nonlinear function is n (as opposed to 3n for unstructured approach), and if value is taken from table, table memory size for single PU would be $16^{10}$, i.e. 40-bit memory address and $1024^4$ byte (at least, now we can name it: 1 Tera byte) memory space. Despite enormous reduction, this value is still high. Different approaches are followed to reduce this value, and approximation of nonlinear functions with neural networks is one of them.

## 3. Cerebellum and CMAC

There is a small number of neuron types present in cerebellum, common to all vertebrates, arranged in highly ordered fashion. Among them are largest neuron in the brain, (giant) Purkinje cell, and most numerous type of neuron, small granule cell [2]. They are organized in three layers. Middle layer, so called Purkinje cell layer, contains only Purkinje cells packed in one-cell thick layer. Inner layer (below Purkinje layer), granule cell layer, inhabits granule cells, Golgi cells, and some other interneurons (some only at specific regions) [15]. Outer layer of cerebellar cortex is named molecular layer. It consists of granule cells axons with characteristic T-shaped form crossing at right angle with plane of flat-shaped Purkinje cells dendrite trees. Present in this layer are also basket and stellate cells. Granule cells are the only excitatory cells in the cerebellum, all the others are inhibitory. Axons of Purkinje cell are sole outputs of the cerebellar cortex, whereas inputs are through group of mossy fibers and through climbing fibers. There are lot of evidences about neuronal connections inside cerebellum, and higher functional organizations, into microzones and even higher to zones, when considering connections with other parts of the brain [16]. Simple sketch of connections between cerebellar neurons is given in Fig. 1. There are two main feedforward routes from input to output. In first main route afferent mossy fibers project to granule cells, granule cells project to Purkinje cell, Purkinje cell sends its efferent axon out of the cerebellar cortex. Second main route is from climbing fiber to Purkinje cell, and again from Purkinje cell out of the cerebellar cortex. Connection climbing fiber-Purkinje cell is rather specific in that each Purkinje cell receives strongest synaptic contact in the brain, and that exclusively from single climbing fiber. Additional routs are through basket and stellate cells inhibitory interneurons. Golgi cell forms local feedback loop with granule cells, and an additional feedforward route from mossy fibers to Purkinje cell output fiber. Connectivity like this is present in all vertebrates, almost invariant, with difference only in quantities [17]. There are also additional similar structures in vertebrates and invertebrates [18]. Other difference is anatomical and has to do with appearance of deep folds that resulted with increase of cerebellar cortex, with purpose of more compact packing. If unfolded human cerebellum would have approximate dimensions of 1 m long and 50 mm wide, with as high as $15 \times 10^6$ Purkinje cells [19]. Number of Purkinje cells in rat is $3.38 \times 10^5$ [20] and in frog around 8300 [21].

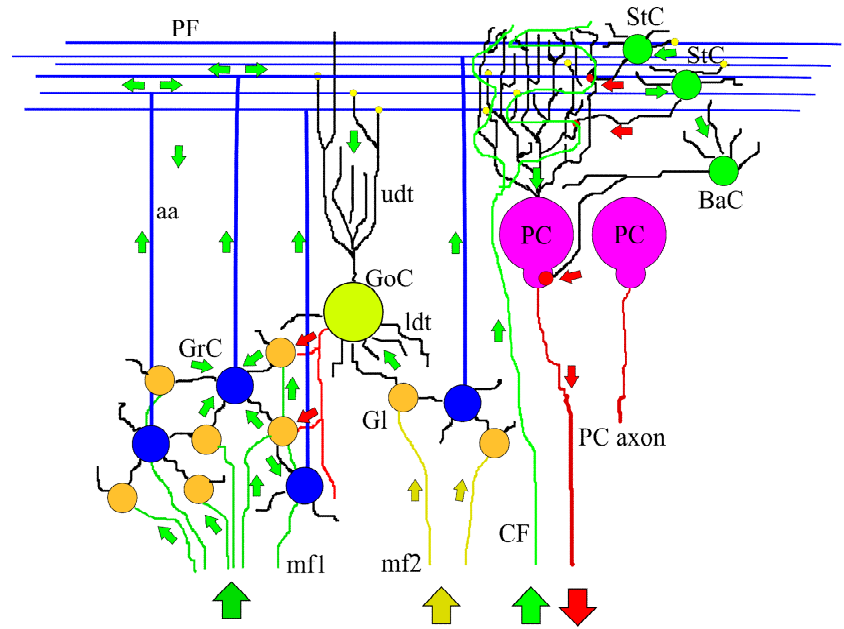

**Fig. 1** Cerebellar neurons, connections, and signal routes. **PC**: Purkinje cell; **GC**: Granule cell; **GoC**: Golgi cell; **StC**: Stellate cell; **BaC**: Basket cell; **Gl**: glomeruli; **aa**: granule cell ascending axon; **PF**: parallel fibers; **adt**: Golgi cell ascending dendritic tree; **ddt**: Golgi cell descending dendritic tree; **mf1**, **mf2**: mossy fibers; **CF**: climbing fiber; **PC axon**: Purkinje cell axon.

Cerebellum is collection of numerous elementary processing units (ePU), thousands to tens of millions, which may, more or less, share processed input data, and influence other ePU with limited number of collaterals from output fiber. To understand function of the cerebellum, as a first step would be to understand function of single ePU by neglecting possible mutual couplings, as they would belong to higher level organization of several ePU. This approach is followed by renowned theories of Marr [22] and Albus [23], and formalized later by Albus with Cerebellar Model Articulation Controller (CMAC) [24] as type of artificial neural network (ANN) that can be used as robotic controller. Model retains two main feedforward routs, with attribution of learning mechanism to climbing fiber-Purkinje cell route. Purkinje cell functions as a perceptron with adjustable weights that model synapses between this cell and granule cell axons (parallel fibers), as only site were learning takes place mediated by climbing fibers assumed to carry error signal. Model does not include explicitly molecular layer inhibitory neurons (basket and stellate cells), but their functionality according to theory is included in possibility of having negative weights. Fig. 2 shows basic CMAC structure. Granule cell layer processing is represented with mapping $M_{IA}$ from input space (mossy fibers) to association space (inputs of the perceptron). This type of ANN is characterized with fast learning, attributed to local generalization property, which in turn is dependent on mapping $M_{IA}$. By selecting proper mappings CMAC can turn to simple perceptron that wont be able to solve even XOR problem, to storage table, to network with properties equivalent to radial basis functions (RBF) ANNs, it can turn to be equivalent with fuzzy ANN. or to some quite different ANN. In case of multidimensional input spaces, standard operation used to process receptive fields is multiplication operation [25] [26] (in standard CMAC with rectangular shaped receptive fields AND operation is used [27]). With this broad range of possibilities, success of the network (perceptron) is highly dependent on proper selection of mapping for selected problem. So, network can learn, but quality and quantity of learning relies on opportunities that are given for the network to learn from associations presented by mapping $M_{IA}$. Since this mapping represents processing in the granular layer, turns attention to process of creating multidimensional receptive fields, and if multiplication really is present in biological neurons [28], [29] [30]. If it is present, is it implemented by a single neuron or by a group of neurons [31], [32]? In other side, coordination of multi-joint systems may have great benefit from it, as dimension of input space can be greatly reduced, because of multiplicative nature of speed and acceleration in dynamics model, Eq. (4). Does multiplication between two different inputs happen in cerebellum? Where possible sites are where multiplication in the cerebellum could take place, and how could they be related to dynamics model computation, as important issue in controlling balance (assisted by evolutionary oldest part of the cerebellum, archicerebellum), and movement coordination (second oldest part of the cerebellum, spinocerebellum or known also as paleocerebellum)?

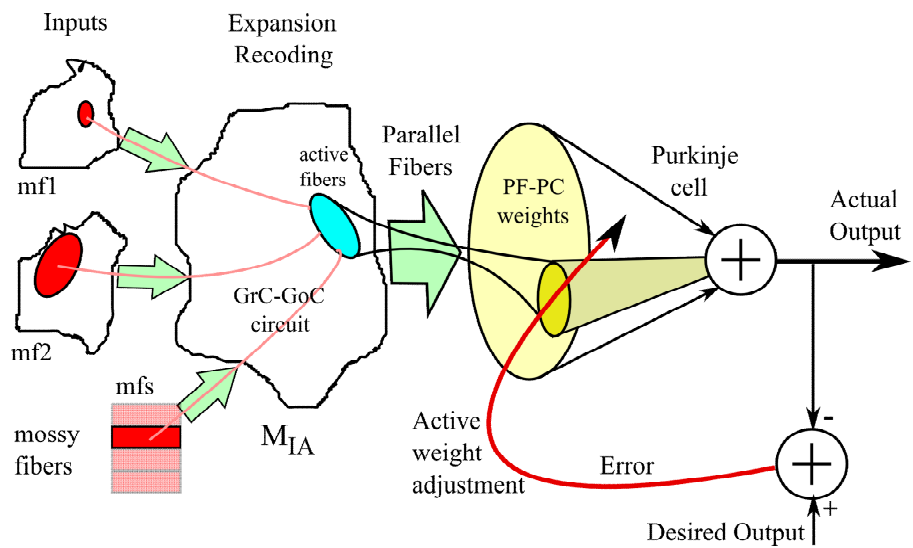

**Fig. 2** CMAC structure.

## 4. New Computational Model of Cerebellar Function

Multi-joint articulated robots dynamics is characterized with many interactions between joints. High quality control requires carefully designed controller, and robot dynamics model is usually part of this controller [33], [34]. With exception of simple cases, finding this model for more complicated structures is a challenge. Having in mind that oldest parts of the cerebellum are related to the problems that in essence contain dynamic interactions, better understanding of cerebellar function will help us to solve robotic problems of ever increasing complexity, which still in number of sensors and actuators is far behind from humans and animals that serve as inspiration [35]. This is new anatomy-based model [36] that gives new functional description for cerebellar circuits, revealing relation to dynamics calculation. Model includes:

- new functional interpretation for granule cell-Golgi cell circuit, and distinction between upper and lower dendrite trees,
- explanation of scenarios for sharing Golgi cells between Purkinje cells,
- new function for basket cells,
- new function for stellate cells, with distinct feature for upper and middle ones,
- definition of coding for different signals at parallel fibers and mossy fibers for consistency with new model,
- direct integration of information from vestibular system,
- direct integration of multilevel interaction with environment,
- explanation of convergence on deep-cerebellar nuclei.

### 4.1 Cerebellar elementary processing unit

Cerebellum will be treated as computer architecture with several levels of organization. Lowest level contains thousands to millions of cerebellar elementary processing units (CePU), with only one output form axon of single Purkinje cell (PC). Functional diagram of CePU is shown in Fig. 3. Most numerous inputs to the PC come from T-shaped axons of granule cells (GrC), know as parallel fibers (PF), and are excitatory by nature. Other inputs to PC are from molecular layer inhibitory interneurons, stellate cells (StC) and Basket cells (BaC). StCs provide input through PC dendrite tree, whereas group of BaCs supply inhibitory synapses to the basket-like structure at initial segment of PC soma. PF provide inputs to the StC and BaC, and serve as data path that is shared between many PCs. Each PC will receive data from as many as 200000 (in humans) from about 10 million PF that will pass by. Data at PF is result of preprocessing done by GrCs and Golgi Cells (GC) over cerebellar input data from mossy fibers (MF).

## 4.2 Granule cells-Golgi cell circuit

Granule cell receive inputs from a number (usually 4-5) of distinct MF and generates output to a single PF. A group of GrC will map information from $m_g$-dimensional MF input space to a sparse $p_g$-dimensional PF output space. Index $g$ is used to indicate existence of many different groups. Marr [22] referred to output of GrC as codon representation of an input, while Albus [23] named mapping process as expansion recoding and output as association of input. In concepts of ANN output would correspond to some sort of multidimensional basis function. Sparsity is considered important feature of new representation and will influence learning speed and capacity.

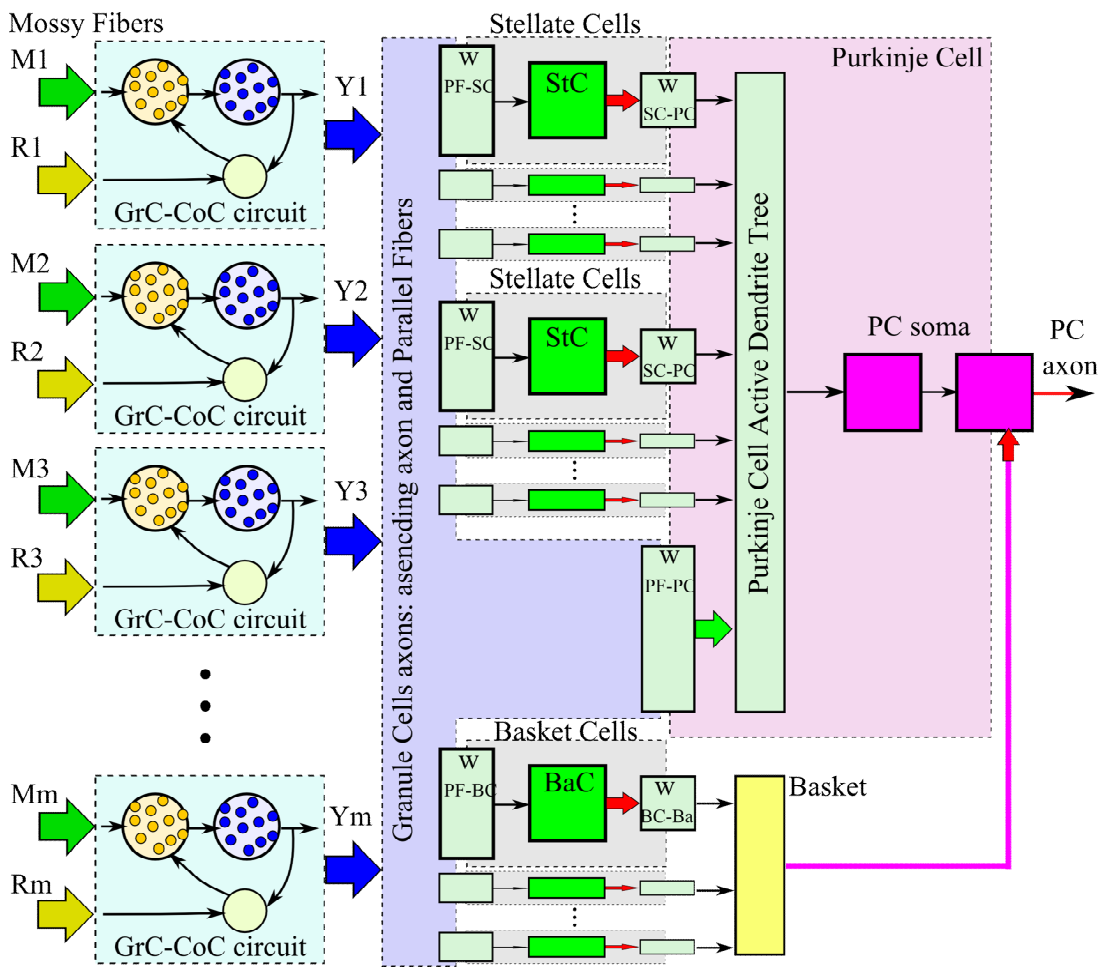

**Fig. 3** Functional diagram of **C**erebellar **e**lementary **P**rocessing **U**nit, CePU.

In an ideal situation, when mossy fibers provide information in form of normalized basis functions, and GrC would function as multiplier of its inputs, output would be normalized higher dimensional basis function. This ideal approach is usually followed when using ANN. GoC are used to deal with real situations and keep mean output activity on controlled level. GoC influences output over two control mechanisms, feedback and feedforward, with former acting through upper part off dendritic tree residing on molecular layer, and synapsing with GrC axons (PF and ascending axon [37]). Feedforward uses lower dendritic tree and synapses with MF. According to Marr [22] output activity is controlled by influencing GrC threshold by that tree that is more powerfully stimulated [22]. Albus suggested that automatic gain control [23] is used to maintain constant output activity independent of input activity. Distinction between two trees is on speed of acting, with feedforward path being faster acting.

Our opinion is that part of cerebellum for handling dynamic interactions (balance control, movement coordination, and joint decoupling) between joints of the multijoint system works in mode of rate coded version of continuous control system. Functional relation between inputs and outputs will be analyzed from functional diagram of GrC-GoC circuit, Fig. 4 (see also Fig. 7 from [23]). GrC receive inputs $L_{jk}$ from structures called mossy fiber rosettes (cerebellar glomeruli); place where GoC axon terminates too and influences mossy fiber signals $M$. Output from single GrC will be

$$Y_i^+ = \left( \sum_{k=1}^{4} M_{ik} - \sigma_i - K_{th}O \right) \cdot G_{Gr} \qquad (9)$$

if control is done by changing GC threshold, and it will be

$$Y_i^+ = \left( g(O) \cdot \sum_{k=1}^{4} M_{ik} - \sigma_i \right) \cdot G_{Gr} \qquad (10)$$

with

$$g(O) = 1 - K_g O \qquad (11)$$

if control is done through automatic gain control. $K_{th}$ and $K_g$ are glomeruli specific constants. Function of GrC is assumed simple summation of four incoming signals from dendrites followed by $G_{Gr}$ part, which may include neuron dynamic and nonlinear static characteristics (except firing threshold $\sigma_i$ already included). Plus sign over $Y$ indicates that value can be only positive or zero, i.e. neurons by nature are excitatory or inhibitory and their functionality can not be reversed. GoC output is

$$O^+ = \left( H_U \cdot \sum_{i=1}^{p} Y_i + H_L \cdot \sum_{j=1}^{q} R_j + \theta \right) \cdot H_{Go} \, . \qquad (12)$$

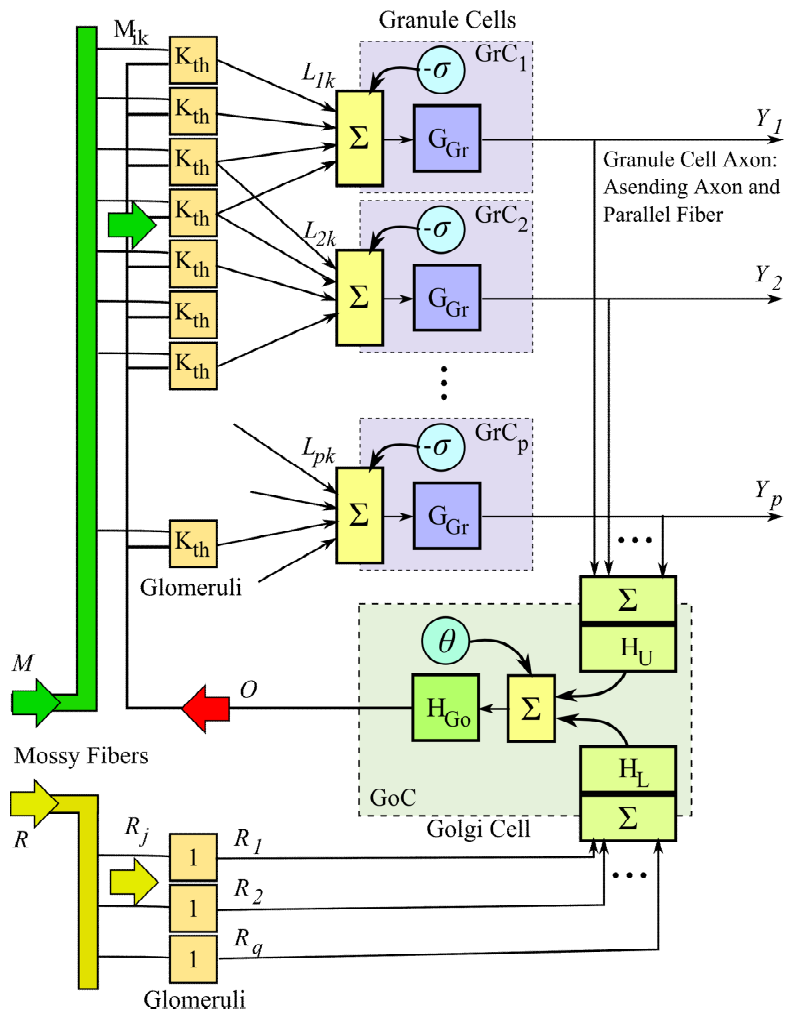

**Fig. 4** Granule cells-Golgi cell functional diagram for preprocessing input data from mossy fibers.

In contrast to [22] it is assumed that action potentials from both dendritic trees are simply summed. Possible behavioral differences are taken into account with functions $H_U$ and $H_L$. $p$ and $q$ denote number of synapses (assumed fixed) made by upper and lower dendritic trees, respectively. Spontaneous firing is represented with $\theta$, while dynamic and static (without spontaneous firing) characteristics of GoC are represented with $G_{Go}$. Plus sign over O, again, has same meaning, whereas minus sign preceding it will indicate inhibitory nature of GoC. Static characteristics of GrC and GoC are given in [38], where it can be seen that GoC is characterized with spontaneous firing activity, while GrC will start firing after crossing some threshold, modeled with $\sigma_i$ and $\theta$, respectively. From Eq. (9), (11), (12) and after summing over $m$ active outputs $Y_i$ we can find

$$\sum_{i=1}^{m} Y_i^+ = \frac{G_{Gr} \cdot \sum_{i=1}^{m} \left( \sum_{k=1}^{4} M_{ik} - \sigma_i - K_{th} H_{Go} \theta \right)}{1 + G_{Gr} \cdot K_g H_U H_{Go} \cdot m} -$$

$$\frac{G_{Gr} \cdot K_g H_U H_{Go} \cdot m \cdot \dfrac{H_L}{H_U} \sum_{j=1}^{q} R_j}{1 + G_{Gr} \cdot K_g H_U H_{Go} \cdot m} \, . \qquad (12)$$

Aim is to keep $m$ in a small fraction of $p$, around 1%. Independence of output activity from input activity $M_{ik}$ cannot be achieved, contrary to conclusion of Albus [23], because only positive term in Eq.

(12) is function of inputs, and by removing that (ex. by making closed loop gain $G_{Gr} \cdot K_g H_U H_{Go} \cdot m$ very large) no output will be active. But despite that, output activity can be kept constant if inputs $M_{ik}$ are in form of basis functions, which would result in constant first term of Eq. (12). Sparsity then is controlled with widths of basis functions. All this could have been done with GrC only, without complicating things with cerebellar glomeruli and GoC. Our conclusion is that usefulness of this structure lies on interpretation of second term of Eq. (12). Since, except input information, all parameters are related to constructive parameters, this equation can be written as

$$Y = k_1 M - k_2 - k_3 R \ . \tag{13}$$

For given M it is equation of a line with negative slope. This slope will effectively turn to positive one at final action: inhibiting inhibitor neuron. Our opinion is that main functionality behind preprocessing with GoC and cerebellar glomeruli is realization of linear dependence of group of outputs $Y_i$ from an additional input $R$, i.e. simple multiplication operation, and preparation phase for approximating function prototypes given by Eq. (5)-(8). Consequence of this functional interpretation is that coding to be used for information on $R_j$ inputs is rate coding with no GoC axon termination on corresponding glomeruli. Outputs $Y_i$ will provide higher dimensional basis functions in M space modulated in amplitude with additional variable R. Same outputs can be used between many PC for final function approximation, ex. when calculating dynamics model terms of form $d_{km}(q)q_m$, same set of $Y_i$ outputs (PF) can be used for all joints, where $M$ would correspond to joint positions vector $q$, and $R$ to acceleration $q_m$. This removes the doubt put by Marr [22] about sharing GoC between PC. One prediction can be made from this interpretation: number of PC that will contact this set of PF should be related to dimension of joint space where dynamic interactions may occur. To calculate whole dynamics equation, for each term in Eq. (4) at least one GoC will be needed. Recoding higher dimensional spaces can be done with several GoC over corresponding subset of dimensions, and number of GrC dendrites could match the dimensionality of subspace. These subsets may have realistic meaning, like shoulder position, head position, etc.

## 4.3 Basket cells and Stellate cells

Role of BaC and StC according to Marr [22] was in setting up the PC threshold, with distinguishing role of most superficial StC as on preventing false initial response by the PC. Albus in his theory [23] assigned them function of providing negative weights for PC that serves a function as a perceptron, a feature that would enable PC high learning capabilities. Many models and theories of cerebellar function do not include them at all [39], or make no difference between them and refer them together as basket/stellate cells [39]. [40] concluded that StC and BaC are similar types of interneurons, with functional difference to synaptic strength, with that of BaC (somatic) being 7-fold higher than that of StC (dendritic). Different spatial, temporal, and physiological effects and consequences for stellate (dendritic), and basket (somatic)-type inhibition was attributed by [39].

When cerebellum is set in context of balance and movement coordination, we hypothesize different function for each of them in duty of computing dynamic interactions. Basket cells are part of the circuit for calculating Coriolis and centrifugal terms of dynamics equation. Stellate cells will mainly handle calculations of friction terms. A set of GrC with one GoC and one PC will be able to approximate calculation of product between one multidimensional nonlinear function and one linearly dependent variable, Eq. (5). This set can be used to calculate terms of inertial torques of Eq. (4)

$$
\begin{aligned}
d_{km}(q)q_m = f_P^k(q, q_m) &= N^k(q)q_m \\
&= \sum_{s=1}^{p} Y_s^{k+} \cdot w_{PCs}^{km} \\
&= \sum_{s=1}^{p} B_s(q)q_m \cdot w_{PCs}^{km}
\end{aligned}
\tag{14}
$$

To calculate Coriolis and centripetal terms an additional multiplication is needed, Eq. (6). One BaC with its axon synapsing on PC soma will approximate this, under assumption that this interaction performs multiplication between these two signals, and $ij$th term for joint $k$ can be written as

$$
\begin{aligned}
h_{kij}(q)q_i \cdot q_j = f_P^k(q, q_i) \cdot q_j &= \left( \sum_{s=1}^{p} Y_s^{k+} \cdot w_{PCs}^{kij} \right) \cdot q_j \\
&= \left( \sum_{s=s_q}^{s_q+m} B_s(q)q_i \cdot w_{PCs}^{kij} \right) \cdot q_j
\end{aligned}
\tag{15}
$$

with $B_s(q)$ being $m+1$ nonzero bases functions for current joint vector $q$, and $w^{kij}_{PCs}$ are weights of PF-PC synapses. This approach would need $n^2$ PC and BaC for single joint. Additionally, PF with joint speed data $q$ is needed. To find more suitable solution we will start from a group of Coriolis and centrifugal terms

$$
h_{ki} = h_{ki1}(q)q_iq_1 + h_{ki2}(q)q_iq_2 + \quad + h_{kin}(q)q_iq_n \, .
\tag{16}
$$

By applying Eq. (15) for given position $q_0$ and square basis functions of width 1, Eq. (16) takes this form

$$
\begin{aligned}
h_{ki} &= B_s(q_0)q_iw_{PCs}^{ki1} \cdot q_1 + B_s(q_0)q_iw_{PCs}^{ki2} \cdot q_2 + \quad + B_s(q_0)q_iw_{PCs}^{kin} \cdot q_n \\
&= B_s(q_0)q_iw_{PCs}^{kii} \left( w_{Xs}^{ki1} \cdot q_1 + w_{Xs}^{ki2} \cdot q_2 + \quad + w_{Xs}^{kin} \cdot q_n \right) \\
&= B_s(q_0)q_iw_{PCs}^{kii} \left( B_s(q_0)q_1w_{BCs}^{ki1} + B_s(q_0)q_2w_{BCs}^{ki2} + \quad + B_s(q_0)q_nw_{BCs}^{kin} \right)
\end{aligned}
\tag{17}
$$

Last form of Eq. (17) can be interpreted as sum of n signals from BaC controls output from one PC. Weights $w^{kij}_{BCs}$ represent synaptic connections PF-BaC. This form as input for BaC uses already present outputs from GrC. Now we need only $n$ PC for single joint, but $n^2$ BaC are needed. To simplify further, BaC that uses basis functions modulated with jth joint speed will be used for all PC that need it, resulting with total Coriolis and centrifugal term for joint k as

$$h_k = B_s(q_0)q_1 w_{PCs}^{k1}\Big(B_s(q_0)q_1 w_{BCs}^{k1} + B_s(q_0)q_2 w_{BCs}^{k2} + \quad + B_s(q_0)q_n w_{BCs}^{kn}\Big) +$$
$$B_s(q_0)q_2 w_{PCs}^{k2}\Big(B_s(q_0)q_1 w_{BCs}^{k1} + B_s(q_0)q_2 w_{BCs}^{k2} + \quad + B_s(q_0)q_n w_{BCs}^{kn}\Big) +$$
$$B_s(q_0)q_k w_{PCs}^{kk}\Big(\quad + B_s(q_0)q_{k-1} w_{BCs}^{k(k-1)} + 0 + B_s(q_0)q_{k+1} w_{BCs}^{k(k+1)} + \quad \Big) + \qquad (18)$$
$$B_s(q_0)q_n w_{PCs}^{kn}\Big(B_s(q_0)q_1 w_{BCs}^{k1} + B_s(q_0)q_2 w_{BCs}^{k2} + \quad + B_s(q_0)q_n w_{BCs}^{kn}\Big)$$

Value zero at $kk^{th}$ term indicates anatomical observations that BaC do not inhibit the PC immediately adjacent [23], that is consistent also with physical interpretation that $h_{kkk}$=0 [7]. Interpretation is valid too for wider basis functions, which will justify much less dense dendritic tree of BaC. Justification is based on the fact that all outputs from a group of GrC modulated with given joint speed carry joint speed information, and it will suffice to sample it sparsely, ex. 2 per basis width $m$, i.e. if $m$=20 number of inputs to BaC would be p/10, meaning ten times less than inputs to GoC, and not necessarily related to number of PC synapses. Even further simplification can be in using fixed PF-BaC synapses. Consequence of this would be that BaC make synapses with ascending axon of corresponding GrC outputs, which are essentially hard wired [41] connections, but there are disagreements about functionality between GrC synapses on ascending axon and PF fiber [42], [43]. Additional possibly adjustable places are BaC-PC synapses, which depending on the configuration may be redundant to GrC-BaC synapses, or the only one adjustable.

Other extreme, in contrast to mammals and birds, is total lack of BaC, like in most fish, amphibians, and most reptiles [39]. This could be interpreted as lack of given type of dynamic interactions, insignificant contribution, or different mechanism of approximation.

There are two friction terms in dynamics equation of motion, dynamic and static. Calculation of dynamic friction is readily achieved with StC that would receive inputs from sparsely sampled joint speed modulated basis functions, like in BaC, where output of the StC would correspond to recovered joint speed information. StC synapse with PC dendrite would represent dynamic friction coefficient $F_d$

$$\tau_{kfd} = F_{kd} q_k = \left(\sum_{s=rnd_1(p)}^{rnd_d(p)} B_s(q) q_k \cdot w_{SCs}^k\right) \cdot w_{SP}^k . \qquad (19)$$

Weights $w_{SCs}^k$ are for PF-StC synapses that may be fixed one, whereas adjustable weight is the one between StC and PC $w_{SP}^k$. Friction effect is local and in ideal case there is one friction term per joint. Since in living organisms many joint are with many DOF driven by many paralleled actuators (muscles), each actuator that drives given joint will need one StC. Those StC at most superficial regions of molecular layer would probably be the ones related to joint with 1-DOF, and deeper one for joints with more DOF, where resultant speed would govern dynamic friction torques.

Calculation of static friction term would need GrC with space encoded joint speed. This task can be accomplished with smaller GoC and small number of GrC. If outputs of these GrC space coded joint speed information, static friction can be calculated as

$$\tau_{kfs} = F_{ks}(q_k) = \left( \sum_{s=1}^{v} B_s(q_k) \cdot w_{SCs}^k \right) \cdot w_{SP}^k .$$  (20)

In this case weights $w_{SCs}^k$ at PF-StC synapses are adjustable, whereas between StC and PC $w_{SP}^k$ is redundant, or can be used as global adjustment parameter.

## 4.4 Functional integration of sensory information

All types of information generated from different that are related to balance, movement, and interaction with environment are easily integrated with this new functional model of the cerebellum. Processed information from vestibular systems (saccule and utricle), combined with information from joints, will result with orientation of body relative to gravitational field. This information can be used in computation of gravity terms of dynamics model in a similar way as was done with inertial torques, Eq. (4) and Eq. (14), with only difference that instead of joint accelerations modulatory functions would be gravity projections.

In a completely same way, and same set of computation prototypes (a set of GrC with one GoC and one PC) can be used to integrate cutaneous mechanoreceptors information for movement and interaction with environment. External forces, torques, and places of action will be found by processing sensor information. Their dynamic effects will be computed with expressions similar to Eq. (14) with modulatory function being six components of interaction force/torque for each segment of multijoint system. Last six terms of Eq. (4) represent this group of interactions, but it shows only one set from many possible, usually robot end-effector force/torque interaction with environment.

## 4.5 Higher levels of organization

Second level of organization between CePU is done with PC collaterals [23], [44]. Third level is constituted from microzones [1], [45], each with a group of around 1000 PC arranged in a narrow longitudinal strip that crosses PF into right angle. PC axons terminate to a group less than 50 deep cerebellar nuclei (DCN) [46]. According to new model, microzone would correspond to computational unit that computes total joint torques, Eq. (4), by collecting together all results from individual CePU. This model can help to resolve dilemma about reason that hundreds of PCs converge to single DCN [47]. Climbing fibers (CF) that innervate PCs of microzone (each CF about 10 PC) come from a group of olivary neurons that tend to be coupled by gap junctions. Also axones of BaC are much longer in longitudinal direction, and stay mainly inside a microzone. It is thought that microzones represent effective cerebellar functional units, and even higher level organization represents functional modules (stripes, zones, and multizonal microcomplexes) [1]. These organizations are marked also with molecular markers, as an additional fact beside anatomical and physiological facts [16].

## 5. Conclusions

New computation model of cerebellar function tries to relate cerebellar neuronal circuits with problem of multijoint dynamics model computations. Multiplicative inclusion of joint speeds, accelerations, gravitational acceleration, and forces/torques of interaction with environment, will make great reduction of dimensionality of problem space to be learned and extend generalization over wide range of multiplicative variables. Model gives functional explanation of connections between main neuron types. To be consistent with model, position information should be space-coded with bases functions of selected width. Same data is additionally amplitude modulated (rate code) with an additional information source. Position information on mossy fibers should be in form of basis functions, whereas other variables that are used multiplicatively, such as joint speed, joint acceleration, etc., use simple rate code. Compared to traditional CMAC, new model may serve as guideline for preprocessing and possibilities of inclusion of pure multiplication in model, that is biologically plausible, for solving problems in robotics with higher level of generalization.

Issues to be resolved in context of new model are functions of Purkinje cell collaterals, and for collaterals of climbing fibers. Additional information for seeking more probable explanation is branching of climbing fiber to 10 Purkinje cells: is it related to redundancy, or 10-fold increase in resolution, or maybe load sharing between paralleled actuators (muscle fibers), or group is related to timing issues like interpolation.